\documentclass[10pt,twocolumn,letterpaper]{article}

\usepackage{wacv}
\usepackage{times}
\usepackage{epsfig}
\usepackage{graphicx}
\usepackage{amsmath}
\usepackage{amssymb}
\usepackage{amssymb,amsthm,amsmath,array}
\usepackage{booktabs}\usepackage[linesnumbered, ruled, vlined]{algorithm2e}

\usepackage[group-separator={,}]{siunitx}
\usepackage{subcaption}
\graphicspath{{imgs/}}
\newcommand\norm[1]{\left\lVert#1\right\rVert}

\usepackage[accsupp]{axessibility} 

\def\wacvPaperID{2000} 

\wacvalgorithmstrack   

\wacvfinalcopy 

\usepackage{color}

\ifwacvfinal
\usepackage[breaklinks=true,bookmarks=false]{hyperref}
\else
\usepackage[pagebackref=true,breaklinks=true,colorlinks,bookmarks=false]{hyperref}
\fi

\begin{document}
	\newcommand{\comm}[1]{}

	\def\titlename{Hardware Aware Evolutionary Neural Architecture Search using Representation Similarity Metric}
	\title{\titlename}
	
	\author{Nilotpal Sinha, Abd El Rahman Shabayek, Anis Kacem, Peyman Rostami, \\Carl Shneider, Djamila Aouada\\
		{\tt\small \{nilotpal.sinha, abdelrahman.shabayek, anis.kacem,peyman.rostami,} \\
		{\tt\small carl.shneider, djamila.aouada\}@uni.lu} \\
		SnT, University of Luxembourg\\
}

\comm{
	\author{Nilotpal Sinha\\
		University of Luxembourg\\
		{\tt\small nilotpal.sinha@uni.lu}
	\and
	Abd El Rahman Shabayek\\
	University of Luxembourg\\
	{\tt\small abdelrahman.shabayek@uni.lu}
	\and
	Anis Kacem\\
	University of Luxembourg\\
	{\tt\small anis.kacem@uni.lu}
	\and
	Peyman Rostami\\ 
	University of Luxembourg\\
	{\tt\small peyman.rostami@uni.lu}
	\and
	Carl Shneider\\
	University of Luxembourg\\
	{\tt\small carl.shneider@uni.lu}
	\and
	Djamila Aouada\\
	University of Luxembourg\\
	{\tt\small djamila.aouada@uni.lu}
}
}

\maketitle
\thispagestyle{empty}
\def\methodname{HW-EvRSNAS}
\begin{abstract}
Hardware-aware Neural Architecture Search (HW-NAS) is a technique used to automatically design the architecture of a neural network for a specific task and target hardware.
However, evaluating the performance of candidate architectures is a key challenge in HW-NAS, as it requires significant computational resources. To address this challenge, we propose an efficient hardware-aware evolution-based NAS approach called HW-EvRSNAS.
Our approach re-frames the neural architecture search problem as finding an architecture with performance similar to that of a reference model for a target hardware, while adhering to a cost constraint for that hardware.
This is achieved through a representation similarity metric known as Representation Mutual Information (RMI) employed as a proxy performance evaluator. It measures the mutual information between the hidden layer representations of a reference model and those of sampled architectures using a single training batch.
We also use a penalty term that penalizes the search process in proportion to how far an architecture's hardware cost is from the desired hardware cost threshold. This resulted in a significantly reduced search time compared to the literature that reached up to $8000\times$ speedups resulting in lower $CO_2$ emissions. The proposed approach is evaluated on two different search spaces while using lower computational resources. Furthermore, our approach is thoroughly examined on six different edge devices under various hardware cost constraints.



\end{abstract}


\section{Introduction}
    
    In recent years, deep learning systems have revolutionized the technology around us across multiple domains such as computer vision~\cite{simonyan2014very, sermanet2013overfeat, zeiler2014visualizing, perez2021detection, garcia2021lspnet, 9620184},
    natural language processing \cite{collobert2011natural, wu2016google, devlin2018bert}, etc.
    These advancements were made possible by the abundance of big data, huge
    growth in computational power, algorithmic improvements, and enhancements in
    hardware acceleration. The proliferation of low-energy Internet of Things (IoT) and edge devices has accelerated technological progress by generating massive amounts of data.
    This has led to the growing need to design deep learning systems that can process such huge amounts of data while consuming limited energy \cite{benmeziane2021comprehensive, Oyedotun_2020_WACV, oyedotun2021deep}.
    However, manually designing highly performant neural networks is very challenging as different tasks require different architectural designs and optimizations.
    Also, the availability of a variety of hardware platforms makes it difficult to design an efficient architecture that performs equally well on all hardware.
    For example, \cite{li2021hwnasbench} showed that for a classification task, the same architecture has different latency values from different edge devices. This led to the need of transitioning from conventional \textit{Neural Architecture Search} (\textit{NAS}) algorithms to more specialized types of algorithms, called \textit{HardWare-aware Neural Architecture Search} (\textit{HW-NAS}) \cite{ijcai2021p592, benmeziane2021comprehensive}.    
    While NAS aims to find the architecture with the best performance for a specific task in a given search space \cite{elsken2018neural}, 
    HW-NAS \cite{ijcai2021p592, benmeziane2021comprehensive} aims to find the architectures with the least trade-off between performance and target hardware usage efficiency.

\begin{figure*}[t]
    \centering
    \begin{subfigure}{0.6\textwidth}
        \includegraphics[width=\linewidth]{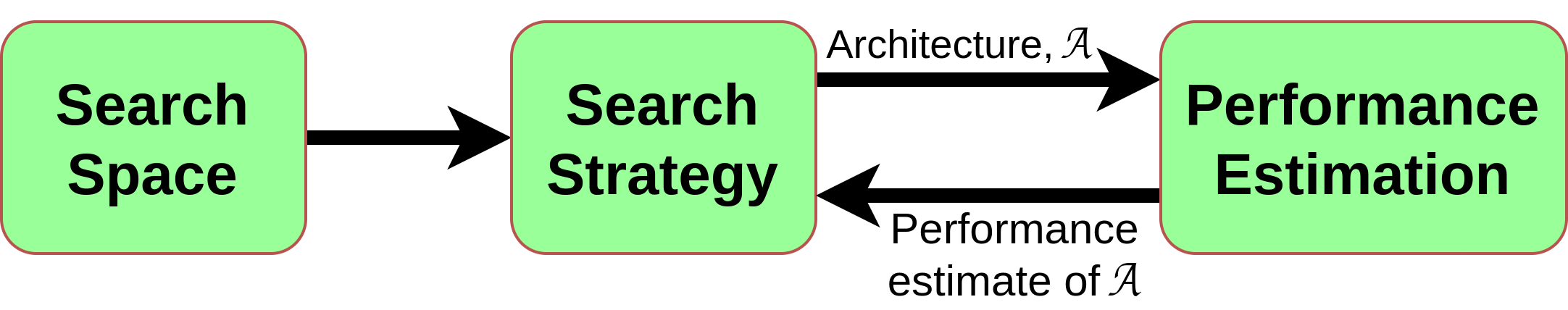}
        \caption{}
        \label{fig:subfig1}
    \end{subfigure}
    \begin{subfigure}{0.70\textwidth}
        \includegraphics[width=\linewidth]{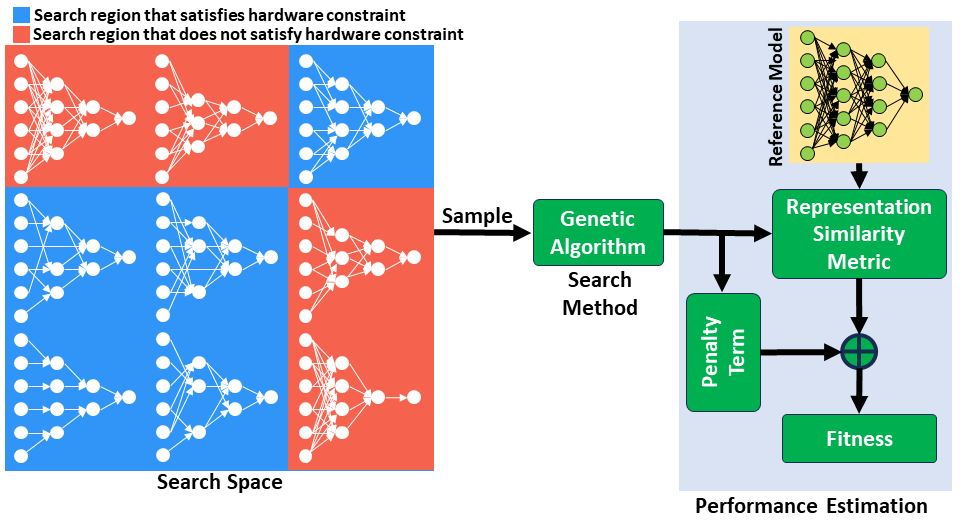}
        \caption{}
        \label{fig:subfig2}
    \end{subfigure}
    \caption{(a) Abstract illustration of Neural Architecture Search methods. (b) Abstract illustration of the proposed method. The blue region in the search space denotes the architectures that satisfy the hardware cost constraint (no penalty) while the red region denotes the architectures that do not satisfy the hardware cost constraint (penalty) for the target hardware.}
    \label{fig:NAS_algorithms}
\end{figure*}
    
    In any NAS algorithm (Figure~\ref{fig:NAS_algorithms} (a)), the estimation of architecture performance is typically the main bottleneck. 
    This estimation is crucial in guiding the search process towards architectures that perform well \cite{elsken2018neural, zoph2017neural, zoph2018learning}.
    As a result, there is a growing demand for HW-NAS algorithms with low search time.
    This is important for reducing the environmental impact, as measured by $CO_2$ emissions.
    Additionally,~\cite{li2021hwnasbench} highlights that as the accuracy of an architecture increases, so too does its latency.
    Therefore, our objective is to develop an efficient HW-NAS algorithm that:
    (1) performs the architecture search with a low search time, and
    (2) takes a desired hardware cost measure (e.g., latency) as input, to identify the best architecture for the target device with a hardware cost lower than the hardware cost metric constraint.     Recent works~\cite{tan2019mnasnet, cai2018proxylessnas, cai2019once} perform the HW-NAS by trying to find the best architecture for a target hardware.
    However, their performance evaluation of the sampled architectures requires a lot of computational resources. 
    To address this issue, we propose an efficient evolution-based HW-NAS method called, \textbf{\methodname}. It reformulates the architecture search problem to find the architecture with the closest performance to a reference model while satisfying a specific target hardware constraint.
    Central to this assumption is that similar performing architectures will have similar Deep Neural Network (DNN) layer representations.
    Recently, \cite{kornblith2019similarity} discussed the desired properties of representation similarity metrics for DNNs.
    A suitable metric should enable finding the closest architecture to a reference model in terms of its representation.
    
    
\comm{
    \textit{Neural architecture search} (NAS) \cite{elsken2019neural} tries
    to replace the reliance on human intuition with an automated search of the neural 
    architecture and have shown promising results in the field of computer vision
    \cite{elsken2018neural}\cite{zoph2016neural}\cite{pmlr-v80-pham18a}\cite{ghiasi2019fpn}.
    Any NAS method \cite{elsken2019neural} has
    three parts (Figure ~\ref{fig:NAS_algorithms}(a)): \textit{search space}, \textit{search
    strategy} and \textit{performance estimation}. The search space typically defines the type
    of architecture that can be represented in principle. The search strategy defines the
    process of how to explore the search space. This typically includes reinforcement learning
    (RL)-based methods \cite{zoph2018learning}\cite{pmlr-v80-pham18a},
    evolutionary algorithm (EA)-based methods \cite{real2019regularized}\cite{sinha2021evolving}
    and gradient-based methods \cite{liu2018darts2}\cite{Zela2020Understanding}. The
    performance estimation refers to the process of estimating the performance of a neural
    architecture. The objective of any NAS method is to find the best architecture with
    high performance using the performance estimation. Such algorithms have also been used to find
    the best architecture for a given task and a given hardware, also known as \textit{hardware-aware
    neural architecture search} (HW-NAS).


     \begin{figure}[h]
		\centering
		\begin{center}
			\includegraphics[width=0.8\linewidth]{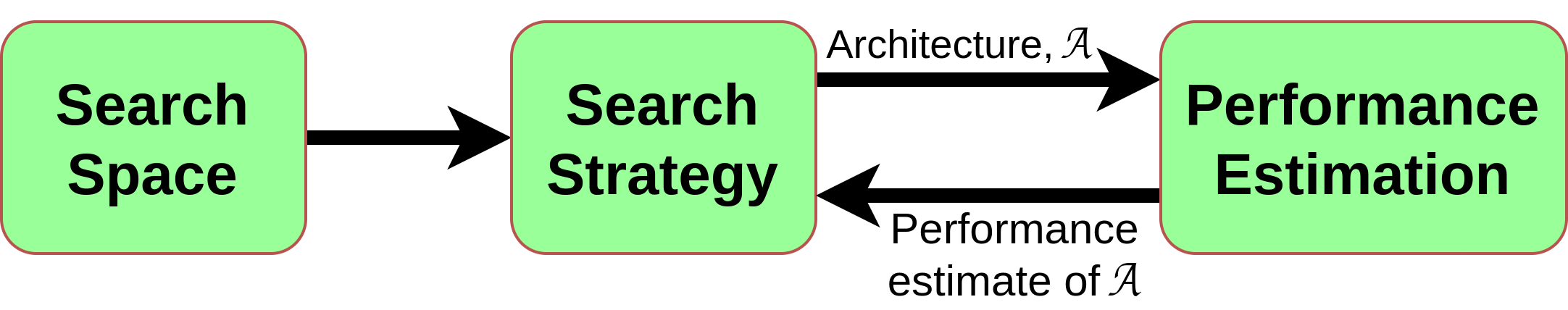}
		\end{center}
		\caption{Abstract illustration of Neural Architecture Search methods.}
		\label{fig:NAS_algorithms}
    \end{figure}
}



    In this work, we sample an architecture from the search space using a genetic algorithm as the evolution-based search method due to its strong performance in NAS problems~\cite{sinha2021evolving, cai2019once}.
    A DNN layer representation similarity metric is used to compute the score between an architecture and a reference model, see Figure~\ref{fig:NAS_algorithms} (b). 
    The \textit{penalty term} block in Figure~\ref{fig:NAS_algorithms} (b) is used to guide the architecture search to the architectures satisfying the hardware constraint.
    The \textit{fitness} (a proxy architecture performance metric) of the sampled architecture is calculated by adding the similarity score and the penalty term.
    The objective of the search method is to converge to the architecture with a high fitness value (more details in Section~\ref{section:proposed_method}).
    To summarize, our main contributions are as follows:
    \begin{itemize}
      \item The HW-NAS is reformulated to find an architecture with similar performance to a reference model for a target hardware while satisfying a specific hardware cost constraint.
      This is achieved by employing a DNN layer representation similarity metric and a penalty term.

      \item The penalty term is designed to guide the search to a sub-space of the whole architecture search space that satisfies a specific hardware constraint.
      In particular, the term penalizes the search process in proportion to how far the hardware cost is from the given hardware cost constraint. We show the effectiveness of the penalty term over the rejection sampling used in previous methods~\cite{cai2019once, zheng2022neural}.

      \item The robustness of the proposed method is demonstrated in two different search spaces for classification tasks and on six different edge devices.

    \end{itemize}
    The rest of the paper is organized as follows: Section~\ref{section:related_work} discusses relevant works.
    The proposed method is explained in Section~\ref{section:proposed_method}.
    Section~\ref{section:experiments} presents the experiments performed to validate the method, 
    and Section~\ref{section:conclusion} concludes the work.
\section{Related Works}
\label{section:related_work}
    Any NAS method, as described in  \cite{elsken2019neural}, consists of three key components (depicted in Figure~\ref{fig:NAS_algorithms} (a)): \textit{search space}, \textit{search strategy}, and \textit{performance estimation}.
    The search space typically defines the types of architectures that can be represented in
    principle. The search strategy defines how the exploration of this search space is conducted.
    This process often involves techniques such as reinforcement learning (RL)-based methods \cite{zoph2018learning, pmlr-v80-pham18a}, evolutionary algorithm (EA)-based methods 
    \cite{real2019regularized, sinha2021evolving, sinha2022neural, sinha2022novelty, 10.1162/evco_a_00331}, and gradient-based methods \cite{liu2018darts2, Zela2020Understanding}.
    
    The performance estimation refers to estimating the performance of a
    neural architecture for a given task.
    This aspect is often the bottleneck of any NAS algorithm.
    In general, it can be categorized in the following ways:
    
    1. \textit{Training from scratch}: In this approach, each architecture is trained from scratch for a certain number of epochs before evaluation.
    However, this method demands significant computational resources \cite{elsken2018neural, zoph2017neural, zoph2018learning}.
    
    2. \textit{Supernet methods}: This approach entails training a single, overly-parameterized network (called a supernet) \cite{liu2018darts2, cai2018proxylessnas} that encompasses all architectures in the search space as its constituent sub-networks.
    Training the supernet only once yields the training of all the architectures within the search space simultaneously.
    While this approach reduces the computational cost, it results in degraded architecture search performance
    as a result of the inaccurate performance estimation by the supernet \cite{bender2018understanding}.
    
    3. \textit{Accuracy predictor based methods}: These approaches employ a smaller model, such as an RNN model \cite{deng2017peephole, liu2018progressive} to predict the accuracy based on the architectural specifications.
    However, training such a model is not straightforward as it requires collecting the dataset of architecture specifications and their corresponding accuracies on the given task, consequently increasing the computational requirements.

Emerging as a novel direction is the evaluation of architecture performance by assessing the similarity of its learned representation to that of a pre-trained, high performing reference model. In this context, Zheng et al. \cite{zheng2022neural} have showcased appealing results for classification tasks by adopting one of the similarity  metrics proposed in \cite{kornblith2019similarity}. 
    Inspired by these promising findings, we employ the DNN layer representation similarity metric utilized in \cite{zheng2022neural} as a fundamental component of our proposed hardware-aware NAS algorithm.

    Hardware-aware NAS algorithms determine the optimum architecture for a target edge device by modifying the performance estimation component in Figure~\ref{fig:NAS_algorithms} (a).
    The assessment of an architecture's performance  is made according to two aspects: (1) accuracy on the given task and (2) its projected computational cost when deployed on the target hardware.
    The objective is to have architectures that exhibit proficient performance (e.g. high classification accuracy) with minimal latency during inference.
    Thus, achieving a favorable hardware cost stands as a pivotal element in any HW-NAS algorithm.
    In this context, a plethora of hardware cost metrics have been employed~{\cite{benmeziane2021comprehensive}}.  These metrics include:
    (1) \textit{FLOPs and Model Size}: In this case, the premise is that the number of parameters and FLOPs correlate positively with model execution time quantified in terms of latency~\cite{smithson2016neural, cai2019once}.
    However, recent research~\cite{li2021hwnasbench, zhang2020fast} has indicated that models with the same FLOPs may indeed exhibit different latencies on different devices.
    The results obtained in our ablation studies (Section~\ref{subsubsection:flopsgood}) corroborate these findings.
    (2) \textit{Latency}: 
    Incorporating actual latency measurements from practical deployment on hardware can enhance the performance of HW-NAS algorithms~\cite{tan2019mnasnet}.
    However, it is important to note that this enhancement comes at the expense of increased search costs.
    Consequently,  many works in the literature resort to employing prediction models~\cite{cai2018proxylessnas, zhang2019neural}, pre-collected look up tables~\cite{wu2019fbnet}, and analytical estimation\cite{zhang2019neural} based techniques.
    In this work, we have designed a HW-NAS algorithm that is flexible enough to accommodate  either FLOPs or latency within the performance estimation block and is agnostic towards the type of the hardware cost metric adopted.

    Since there are multiple objectives in HW-NAS, existing methods generally tackle this challenge by pursuing two distinct strategies:
    \cite{benmeziane2021comprehensive}:
    (1) \textit{Multi-Objective Optimization}: Striking a balance between the objective of obtaining the finest accuracy architecture and the goal of minimizing hardware costs involves trade-offs as the two are inherently conflicting.
    The core challenge lies in identifying pareto-optimal solutions~\cite{ying2020neural, chu2020multi}. Notably, Pareto optimal solutions refer to solutions that cannot be enhanced in one objective without sacrificing at least one other objective. For example, enhancing the accuracy of an architecture might involve increasing network parameters, consequently raising the hardware cost.
    Finally, it is worth noting that this strategy does not grant us control over the desired latency of an architecture for a target device.
    (2) \textit{Single-Objective Optimization}: In this strategy, hardware cost is regarded as a constraint presented in the form of thresholds to be respected during the search process.
    A method like \cite{cai2019once} employs \textit{rejection sampling} to rule out any architecture that does not
    satisfy the constraint during the search process.
    However, rejection sampling suffers from the risk of the halting problem when it rejects all architectures for not satisfying an excessively low hardware cost constraint (discussed in Section~\ref{subsubsection:penalty}).
    In contrast, we use a penalty term that reduces the performance metric of the architecture with respect to its proximity  to the constraint threshold.
    In a similar work, \cite{tan2019mnasnet} also employs a penalty term for the hardware aware part, but their approach introduces two extra hyper-parameters in the penalty term which requires additional  efforts for finding the optimal values for those hyper-parameters, whereas in our method, no extra hyper-parameters are introduced in the penalty term.
    \comm{
    Methods like \cite{tan2019mnasnet}\cite{cai2018proxylessnas} also use penalty term for the hardware
    cost but they do not have precise control over the exact hardware cost value. In contrast,
    in the proposed method, the user specifies the exact maximum hardware cost (latency or FLOPs)
    allowed for the discovered architecture.
}

\section{Proposed Method}
    \label{section:proposed_method}
    \comm{
    As shown in Figure~\ref{fig:NAS_algorithms} (a), the performance estimation is used to validate
    the quality of an architecture. Any NAS algorithm uses it to find the best architecture in the
    search space. A naive way to evaluate an architecture will be to train each architecture
    from scratch for certain epochs and then evaluate its performance on the dataset for the
    given task. However, this approach requires a lot of computational resources (e.g. 3150 GPU 
    days\cite{real2019regularized}) which makes it infeasible.
}
    As explained in Figure~\ref{fig:NAS_algorithms} (b), the architecture search problem is stated as a search method whose objective is the following:
    (1) Find an architecture in the search space with similar performance with respect to a baseline model called \textit{reference model}, measured in terms of a performance metric score for a given task.
    (2) The searched architecture needs to satisfy the hardware cost constraints for the target hardware.
    We use a representation similarity metric called Representation Mutual Information (RMI) \cite{zheng2022neural, kornblith2019similarity} as it has shown promising results as an efficient performance estimation method. Formally, given a pre-trained \textit{reference model} $\alpha^*$ with desired performance metric (e.g. classification accuracy for classification task), an architecture search space  $\mathcal{A}$, a device with a hardware cost constraint $\Omega$, \methodname~ performs the architecture search in the given search space such that the discovered architecture has high performance metric score while satisfying the hardware constraint as follows, 
    
     \begin{equation}
	\label{eq:problem}
        \max_{\alpha \in \mathcal{A}} \;\phi(\alpha^*, \alpha), \quad s.t. \: \Psi(\alpha) < \Omega \ , 
    \end{equation}
\noindent where $\alpha$ is an architecture in the search space $\mathcal{A}$, $\Psi(.)$ is the function that measures the hardware cost (e.g. FLOPs, latency etc.), and $\phi(\alpha^*, \alpha)$ is the deep neural network representation similarity metric that measures the representation similarity between $\alpha^*$ and $\alpha$, which is used as a proxy architecture performance estimator.

\subsection{Performance Estimator}
\label{subsect:performance}
    In order to find the closest architecture with similar representation with respect to the reference model,
    \cite{zheng2022neural} used one of the proposed representation similarity metrics in \cite{kornblith2019similarity} called RMI.
    It measures the mutual information between hidden layer representations of an architecture and the hidden layer representations of the reference model.
    RMI score is calculated using a single training batch which makes it a faster and efficient alternative to the naive computationally intensive train from scratch strategy.
    In particular, given a pre-trained reference architecture, $\alpha^*$, with a specific performance metric for a task (i.e. \textit{reference model}), the RMI score of any architecture, $\alpha$, sampled from the search space is defined as, 
    \vspace{-0.1in}    
    \begin{equation}
	\label{eq:RMI}
        \phi(\alpha^*, \alpha) 
        = \sum_{i=1}^{L}
        \frac{\norm{X^{i*^{T}} X^{i}}^{2}_{F}}{\norm{X^{i*^{T}} X^{i*}}_{F} \norm{X^{i^{T}} X^{i}}_{F}} \ , 
    \end{equation}
    
    \noindent where $X^{1*}, X^{2*}, .., X^{L*}$ and $X^{1}, X^{2}, .., X^{L}$ represent the random variables of feature maps in each layer of $\alpha^*$ and $\alpha$, respectively, and $\norm{.}_{F}$ is the Frobenius norm.
    In order to use the RMI score as performance estimator for the proposed \methodname, each sampled architecture $\alpha$ is first trained on a single selected batch using the following loss function, 
    
    
    \begin{equation}
	\label{eq:loss}
        \mathcal{L}_{loss} = \beta \: \phi(\alpha^*, \alpha) + (1-\beta) \: \mathcal{L}_{task} \ , 
    \end{equation}

    \noindent where $\mathcal{L}_{task}$ is the task specific loss term (e.g. classification loss for classification task), $\phi(\alpha^*, \alpha)$ is the RMI loss term, and $\beta$ is a parameter weighting the contribution of the two loss terms. Once the sampled architecture $\alpha$ is trained, the RMI score defined in Eq.(\ref{eq:RMI}) is used as its performance metric.
    

    \begin{algorithm}[t]
		\caption{\methodname}	
		\label{algo}
		\SetAlgoLined
		\KwIn{Reference model $\alpha*$, Search space $\mathcal{A}$, Hardware constraint $\Omega$,
            Total generations $N_{gen}$, Population size $N_{pop}$, training epochs $N_{train}$}
		\KwOut{Searched architecture, $\alpha^{(N_{gen})}$}
		$g \gets 0$ (Initialize the generation counter)\;
            Initialize the population $\mathcal{P}$ with random architectures\;		
		\While{ $g \leq N_{gen}$ }{
			\For{each individual architecture ($\alpha$) in $\mathcal{P}$} {
				Train $\alpha$ using Eq.(\ref{eq:loss}) for $N_{train}$ epochs\;
                    Evaluate its fitness score using Eq.(\ref{eq:fitness}) \;
			}   
                $\alpha^{(g)} \gets \:$Best architecture in $\mathcal{P}$ \;
                $\mathcal{P} \gets$ Create next generation population using crossover and mutation\;
			$g \gets g + 1$\;
		}		
    \end{algorithm}

\subsection{Hardware Aware Architecture Search}
    As shown in Figure~\ref{fig:NAS_algorithms}~(b), \methodname$\ $uses Genetic Algorithm (\textit{GA})~\cite{sinha2021evolving} for performing the architecture search in the given search space.
    GA begins with a population of architectures and each architecture, $\alpha$, in the population is given a fitness value
    
    \vspace{-0.2in}
    \begin{equation}
	\label{eq:fitness}
        fitness(\alpha^*, \alpha, \Omega) = \phi(\alpha^*, \alpha) + \psi(\alpha, \Omega) \ ,
    \end{equation}

    \noindent where $\psi(\alpha, \Omega)$ is the penalty term which is used for guiding the search method towards regions in the search space that satisfy the hardware constraint. The fitness of an architecture is a combination of performance metric for the given task and a penalty term.
   The goal of GA is to generate/evolve the next generation population such that the fitness values of the architectures in the population increases (i.e. \textit{maximizing the fitness}).
    For a given hardware cost constraint, $\Omega$, the penalty term, $\psi$, for an architecture, $\alpha$, is defined as follows, 
    \vspace{-0.1in}
    \begin{equation}
	\label{eq:penalty}
        \psi(\alpha, \Omega) = 
        \begin{cases}
          \qquad \quad\:\:\:\: 0,  & \Psi(\alpha) \leq \Omega \\
          \:\Omega - \Psi(\alpha), & \Psi(\alpha) > \Omega
        \end{cases}
    \end{equation}
    
    \noindent where $\psi(\alpha, \Omega)$ penalizes the search method by reducing the fitness value of $\alpha$ if it does not satisfy the hardware cost constraint $\Omega$.
    Note that the penalty function penalizes the search method in proportion the proximity of hardware cost to the given hardware constraint.
    In other words, the closer a penalized architecture is to the hardware constraint, the lower is the penalty value.
    For example in Figure~\ref{fig:NAS_algorithms}~(b), the penalty term for sampled architectures that satisfy the hardware constraint (blue region) is $0$.
    Whereas the penalty term value for those that do not satisfy the hardware constraint (red region) is negative. This in turn reduces the fitness value defined in Eq.(\ref{eq:fitness}).

\begin{table*}[h]
    \caption{Comparison of the proposed method with other NAS methods in OFA-search-space for ImageNet dataset and FLOPs (MACs) as a hardware cost constraint.
    “\textit{Manual}” and “\textit{Auto}” in  “\textit{Method Type}” refer to hand-crafted and NAS methods respectively.
    “$CO_2$e” denotes the $CO_2$ emission which is calculated based on~\cite{strubell-etal-2019-energy}. $^\dagger$ indicates fine-tuning to make a fair comparison with our settings.}
    \label{table:OFA}
    \centering
    \begin{tabular}{l|c|c|c|c|c}
    \hline
    Model&Method&ImageNet& MACs & Search Cost & $CO_2$e \\
    & Type &Top1(\%) & & (GPU Hours) & (lbs) \\
    \hline
    Mobilenetv2 \cite{sandler2018mobilenetv2} & Manual & 72.0 & $300$M & 0 & - \\
    ShuffleNet \cite{zhang2018shufflenet} & Manual & 71.5 & $292$M & 0 & - \\
    ShuffleNetV2 \cite{ma2018shufflenet} & Manual & 72.6 & $299$M & 0 & - \\
    \hline
    NASNet-A \cite{zoph2018learning} & Auto & 74.0 & $564$M & \num[group-separator={,}]{48000} & $13.6$k \\
    DARTS \cite{liu2018darts2} & Auto & 73.1 & $595$M & $96$ & $27.4$ \\
    MnasNet \cite{tan2019mnasnet} & Auto & 74.0 & $317$M & \num[group-separator={,}]{40000} & $113.4$k \\
    FBNet-C \cite{wu2019fbnet} & Auto & 74.9 & $375$M & $216$ & $61$ \\
    ProxylessNAS \cite{cai2018proxylessnas} & Auto & 74.6 & $320$M & $200$ & $57$ \\
    SinglePathNAS \cite{guo2020single} & Auto & 74.7 & $328$M & $312$ & $88.1$ \\
    AutoSlim \cite{yu2019autoslim} & Auto & 74.2 & $305$M & $180$ & $51$ \\
    
    OFA$^\dagger$\cite{cai2019once} & Auto & 74.7 & $270$M & $40$ & $11.3$ \\
    \hline
    \methodname$\ $(Ours) & Auto & 74.6 & $240$M & $5$ & $1.05$ \\
    \methodname$\ $(Ours) & Auto & 75 & $300$M & $5$ & $1.05$ \\
    \methodname$\ $(Ours) & Auto & 76.8 & $374$M & $5$ & $1.05$ \\
    
    \end{tabular}
\end{table*}

    The pseudo-code of \methodname$\ $is given in Algorithm~\ref{algo}.
    The method begins with a population of architecture, $\mathcal{P}$, with $N_{pop}$ (population size) number of random architectures sampled from $\mathcal{A}$.
    \methodname$\ $runs for $N_{gen}$ generations.
    In each generation, each architecture $\alpha$ in the current generation population is evaluated by training $\alpha$ using the loss function in Eq.(\ref{eq:loss}) for $N_{train}$ epochs.
    Note that Eq.(\ref{eq:loss}) requires only single training batch and hence will be significantly faster as compared to normal training using only ${L}_{task}$ on all the training batches.
    After training, the fitness of the architecture $\alpha$ is evaluated.
    Once the fitness value is computed, \methodname$\ $creates the next generation population using crossover and mutation~\cite{sinha2021evolving} with the goal of maximizing the fitness value.
    The best architecture, $\alpha^{(N_{gen})}$, after $N_{gen}$ generations is returned as the discovered architecture while satisfying the hardware constraint, $\Omega$.
    Note that the best architecture refers to the architecture in the population with the highest fitness value for a given generation.

\section{Experiments}
\label{section:experiments}
    Details about the experiments such as search space are provided in Section~\ref{subsection:search_space}, implementation details in Section~\ref{subsection:implementation} and datasets in supplementary.
    The re-phrasing of the HW-NAS problem allows us to perform the architecture search with less computational resources.
    To illustrate this, the architecture search is demonstrated on the classification task in order to compare our method to the literature in Section~\ref{subsection:results}.
    The architecture search performance is reported for six different edge devices under different hardware cost settings for those devices in Section~\ref{subsection:results}.
    Finaly, ablation studies to shed light on the effects of various design choices are reported in Section~\ref{subsection:ablation}.
    The code is available at \href{https://gitlab.uni.lu/cvi2/elite/hw-evrsnas}{https://gitlab.uni.lu/cvi2/elite/hw-evrsnas}

\subsection{Search Space}
\label{subsection:search_space}
    The effectiveness of the proposed method is demonstrated on two search spaces:
    (1) \textbf{OFA-search-space}~\cite{cai2019once}: Here, the type of operation (e.g. convolution,  max pooling, etc) are fixed.
    The HW-NAS algorithm aims to find the optimal setting of the model width, depth, and its kernels sizes. FLOPs are used as a hardware cost metric (constraint) to compare \methodname$\ $ with the existing literature on HW-NAS, see Table~\ref{table:OFA}.
    (2) \textbf{NAS-Bench-201}~\cite{Dong2020NAS-Bench-201} provides a unified benchmark for almost any up-to-date NAS algorithm by providing the results on CIFAR-10, CIFAR-100 and ImageNet16-120.
    The objective is to search for the type of the operation (i.e. convolution 3x3, convolution 1x1, max pooling 3x3, skip connect and none) present between two nodes.
    Note that \textit{none} operation is used for denoting that there is no connection between two nodes.
    However, the benchmark does not provide the hardware-cost of the architectures in the search space.
    For this, we use \textbf{HW-NAS-Bench} \cite{li2021hwnasbench} which augments NAS-Bench-201 by providing various
    hardware-cost of all the architectures in the search space for six edge devices: \textit{NVIDIA Edge
    GPU Jetson TX2, Raspberry Pi 4, Edge TPU, Pixel 3, ASIC-Eyeriss, and FPGA}.

\comm{
\subsection{Datasets}
\label{subsection:datasets}
    \textbf{CIFAR-10} and \textbf{CIFAR-100} \cite{krizhevsky2009learning} have 50K training images and 10K testing images with images classified into 10 and 100 classes respectively.
    \textbf{ImageNet}\cite{imagenet_cvpr09} is a well known benchmark for image classification containing 1K classes with 1.28 million training images and 50K test images.
    \textbf{ImageNet-16-120} \cite{chrabaszcz2017downsampled} is a variant of ImageNet which is downsampled to 16x16 pixels with labels $\in\left[0,120\right]$ to construct ImageNet-16-120 dataset.
}

\begin{figure*}[t]
    \centering
    \begin{center}
        \includegraphics[width=\linewidth]{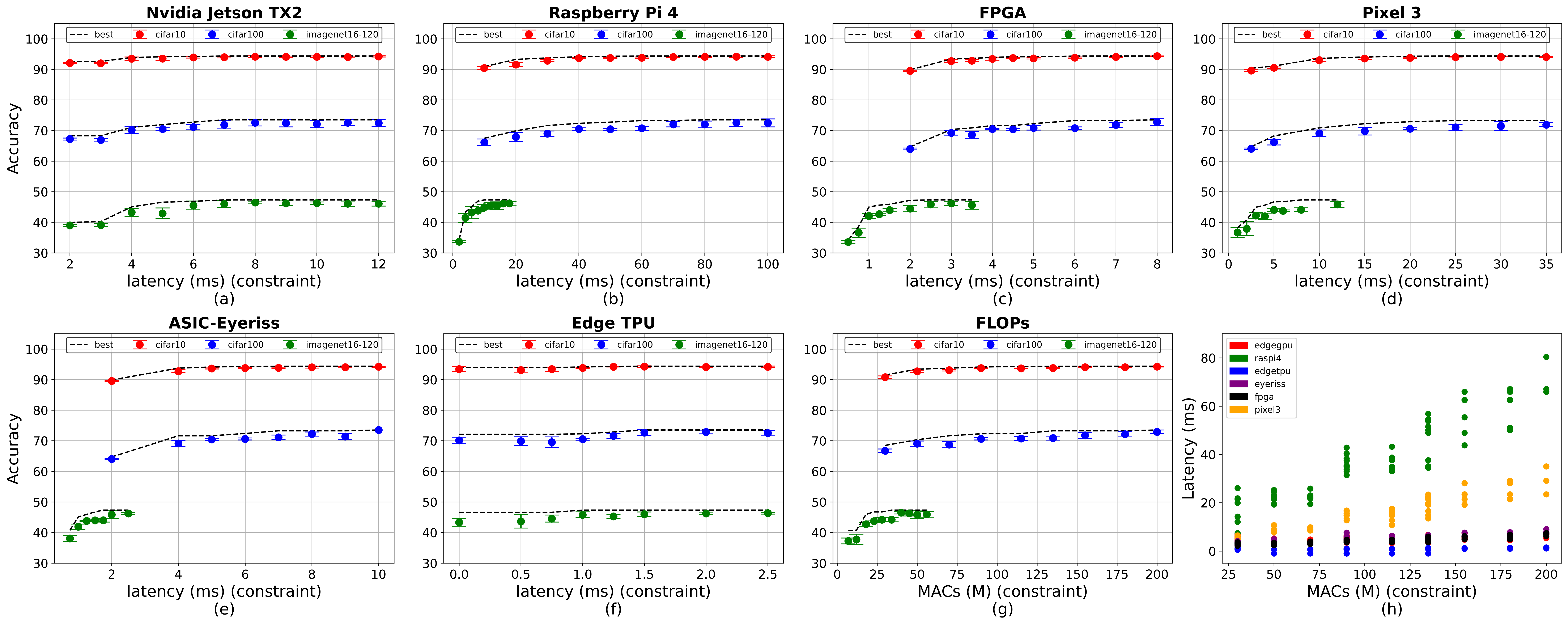}
    \end{center}
    \caption{
    (a)-(g) shows the performance (test accuracy in y-axis) of \methodname$\ $with NVIDIA Jetson TX2, Raspberry Pi 4, FPGA, Pixel 3, Edge TPU, ASIC-Eyeriss, TPU latency constraint, flops value constraint respectively in the x-axis.
    The \textit{dashed} line shows the best architecture under the specific constraint.
    The different colored dots represents the mean$\pm$std accuracy of 10 experiments with different random seeds for the specific constraint.
    (h) shows the latency of the searched architecture (y-axis) for different edge devices when MACs/FLOPs is used as hardware constraint (x-axis).
    }
    \label{fig:exp_all_edge_devices}
\end{figure*}

\subsection{Implementation Details}
\label{subsection:implementation}
    Since both search spaces involve searching over different aspects of the neural architecture, different architecture representations are used for them.
    For the OFA-search-space, the architecture representation used in \cite{cai2019once} is employed and for NAS-Bench-201, we use the architecture representation used in \cite{sinha2021evolving}.
    Architecture search for both search spaces are done using a single NVIDIA RTX A4000 GPU with a population size ($N_{pop}$) of 20.
    For the OFA-search-space, the architecture with highest values for model width, depth and kernel sizes is used as a reference model.
    For NAS-Bench-201, ResNet-20 is used as the reference model following \cite{zheng2022neural}.
    The RMI score for both search spaces are calculated after training for 100 epochs ($N_{train}$) using a value of 0.8 for $\beta$ in Eq.(\ref{eq:loss}).
    The architecture search is performed for 100 generations ($N_{gen}$) for both search spaces.
    
\begin{figure*}[h]
    \centering
    \begin{center}
        \includegraphics[width=\linewidth]{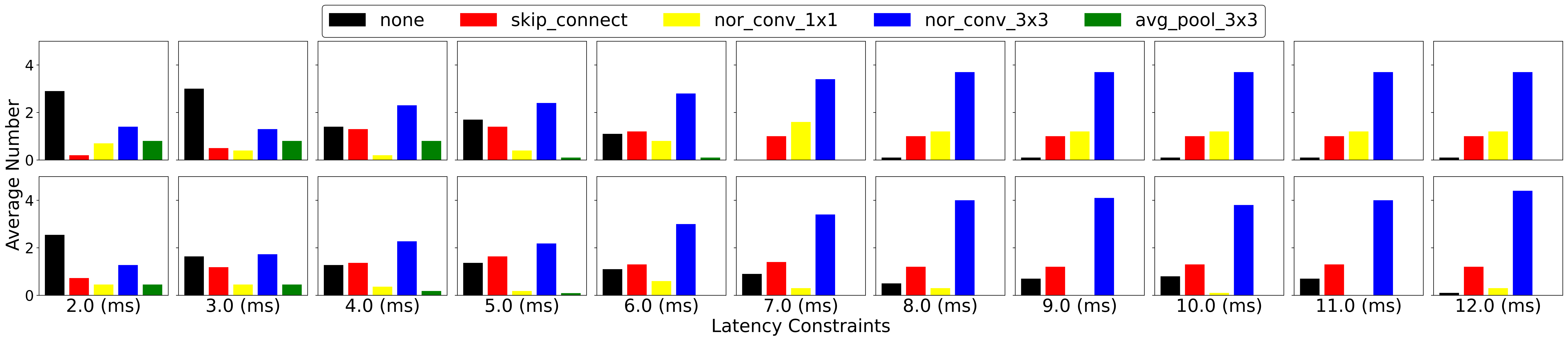}
    \end{center}
    \caption{y-axis represents the average number of 5 different types of operations from NAS-Bench-201 search space.
    Each column of subplot presents the result for different latency constraint.
    \textit{Top row}: Average number of different types of operations present in the top-10 architectures in terms of test accuracy on CIFAR10 dataset on NVIDIA Jetson TX2.
    \textit{Bottom row}: Average number of different types of operations present in the 10 architectures discovered from 10 different search runs for a given latency constraint.}
    \label{fig:c10_ops_dist_edgegpu_latency}
\end{figure*}

\subsection{Results}
\label{subsection:results}
    The results of \methodname$\ $on OFA-search-space for different hardware constraints are shown in Table~\ref{table:OFA} where FLOPs are used as the hardware cost constraint.
    The table shows that the proposed method is able to find architectures in the Top1\% classification accuracy with a low search cost. This cost is the number of GPU hours required to perform the architecture search.
     The lower the cost, the more efficient the NAS method is. It is considered more environmental friendly if it has a lower carbon footprint (measured as $CO_2$e (lbs). Table~\ref{table:OFA} shows that \methodname$\ $discovers better performing architectures than FBNet~\cite{wu2019fbnet}, ProxylessNAS~\cite{cai2018proxylessnas}, SinglePathNAS~\cite{guo2020single}, and AutoSlim~\cite{yu2019autoslim} while having lower FLOPs and using lower search cost (i.e. computational resources) which in turn has a lower carbon footprint.
    In particular, \methodname$\ $performs the HW-NAS in a significantly lower search time: \textbf{8000 $\times$ faster} than MnasNet~\cite{tan2019mnasnet}, \textbf{40 $\times$ faster} than ProxylessNAS~\cite{cai2018proxylessnas} and \textbf{8 $\times$ faster} than OFA~\cite{cai2019once}.
    Note that the reported GPU hours in Table~\ref{table:OFA} are taken from the respective papers as most of them use different GPU for their experiments.
    
    The proposed method is agnostic to the search space, utilized device and hardware constraint. Results of \methodname$\ $ applied on NAS-Bench-201 are reported in Figure~\ref{fig:exp_all_edge_devices} (a)-(f) for six different edge devices. Hardware latency is taken as the hardware cost constraint. The architecture search results using FLOPs as hardware constraint are presented in Figure~\ref{fig:exp_all_edge_devices} (g).
    Note that Figure~\ref{fig:exp_all_edge_devices} (a)-(g) shows the mean and standard deviation of 10 experiments performed with different random seeds for each hardware constraint.
    The x-axis represents the different hardware cost constraint values used for architecture search for a specific hardware device.
    For example in Figure~\ref{fig:exp_all_edge_devices} (a), entries for 4 ms (x-axis) represent the architecture search results for the Nvidia Jetson TX2 with 4 ms latency constraint for three different datasets.
    The figure shows that \methodname$\ $is able to find the closest architecture to the optimal one (shown as dashed line) under different latency constraints for the edge devices.
    It is important to note that the proposed method performance will depend on how accurate the hardware cost metric is, as will be discussed in Section~\ref{subsection:ablation}.

    The proposed method is also investigated on its ability to capture the distribution of operations of top-10 performing architectures for a target hardware at different latency constraints.
    Note that this can be viewed as the sub-space with highest quality architectures under the given latency constraint.
    Figure~\ref{fig:c10_ops_dist_edgegpu_latency} (Top) shows that the average number of \textit{none} operation 
    is more for the latency constraint of 2.0s. As the latency increases, the number of \textit{nor\_conv\_3x3} operation (convolution operation with 3x3 kernel size) increases in the top-10 architectures for the classification task on CIFAR10 on NVIDIA Jetson TX2. This behaviour reflects the harmony between the imposed latency constraint and the retrieved architectures.
    Figure~\ref{fig:c10_ops_dist_edgegpu_latency} (Bottom) plots the average number of different operations present in the architectures discovered in 10 independent runs of \methodname$\ $for each latency constraint. It shows that \methodname$\ $is able to capture architectures with similar operations distribution.  
    In other words, \methodname$\ $ is able to converge to the sub-space with highest performing architectures given certain constraint. This similarity pattern is observed for other edge devices; Raspberry Pi 4, Edge TPU, Pixel 3, ASIC-Eyeriss, FPGA on all the datasets, see the supplementary material.

\begin{figure}[h]
    \centering
    \begin{center}
        \includegraphics[width=0.85\linewidth]{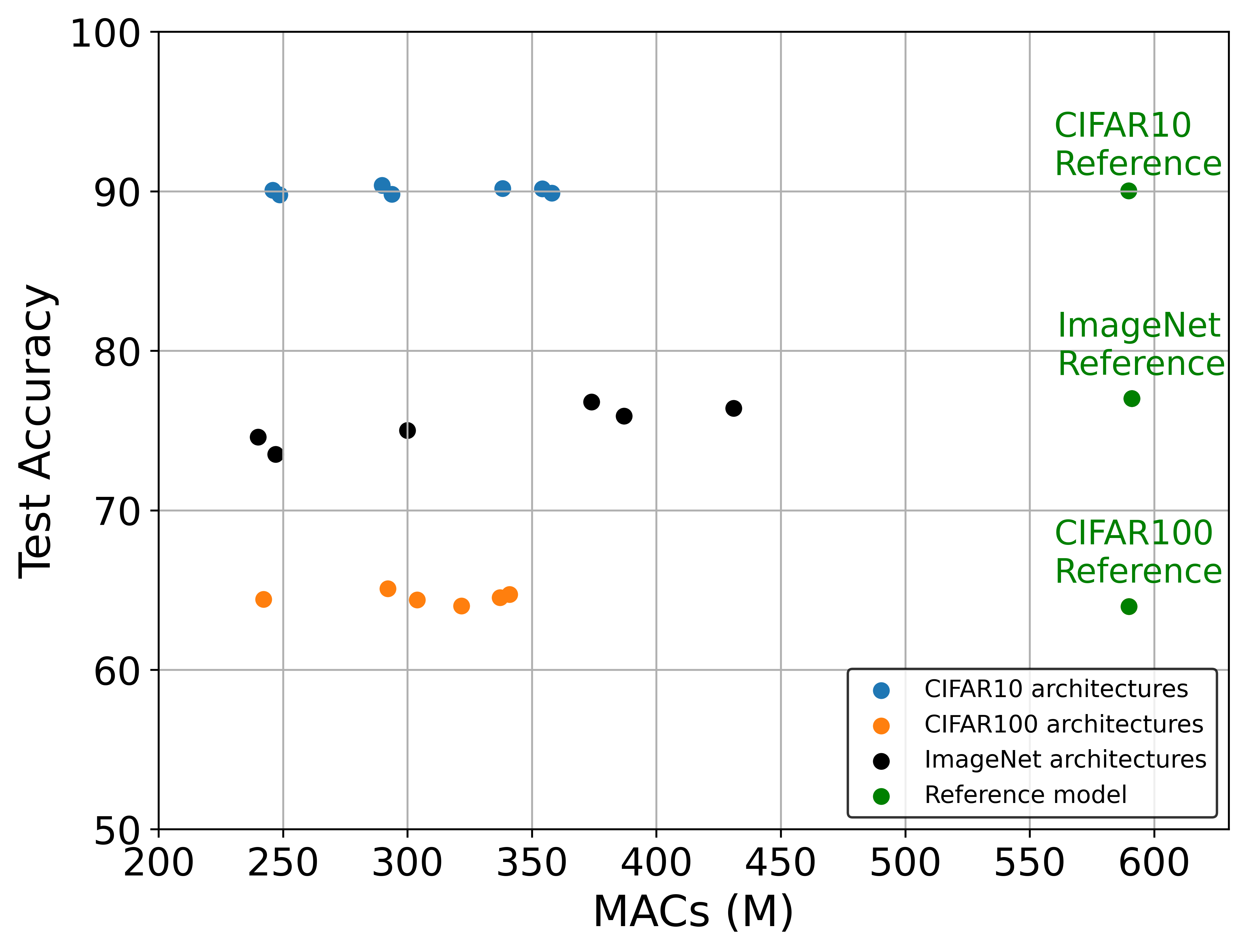}
    \end{center}
    \caption{Effect of the use of DNN layer representation similarity metric in HW-NAS (y-axis represents test accuracy) with FLOPs (MACs) as the hardware cost constraint (x-axis).}
    \label{fig:ofa-ablation-how-works}
\end{figure}
\vspace{-0.2in}
\subsection{Ablation Study}
\label{subsection:ablation}
\subsubsection{Effect of Rephrasing HW-NAS Problem}
\label{subsubsection:RMIeffect}
    The formulation of RMI score, Eq.(\ref{eq:RMI}), measures the mutual information between layers of a reference model and architectures in the search space. Hence, a search looking for an architecture with maximum RMI score under a certain constraint finds one whose layers have the maximum mutual information with that reference model.
    Figure~\ref{fig:ofa-ablation-how-works} shows the test accuracy of the discovered architectures and the reference model in OFA-search-space for three different datasets (CIFAR10, CIFAR100 and ImageNet). 
    It finds architectures with similar accuracy to the reference model while satisfying the FLOPs constraint. Hence, the rephrased formulation of the HW-NAS problem has successfully retrieved highly constrained and efficient architectures for a given a reference model.

\begin{figure}[t]
    \centering
    \begin{center}
        \includegraphics[width=\linewidth]{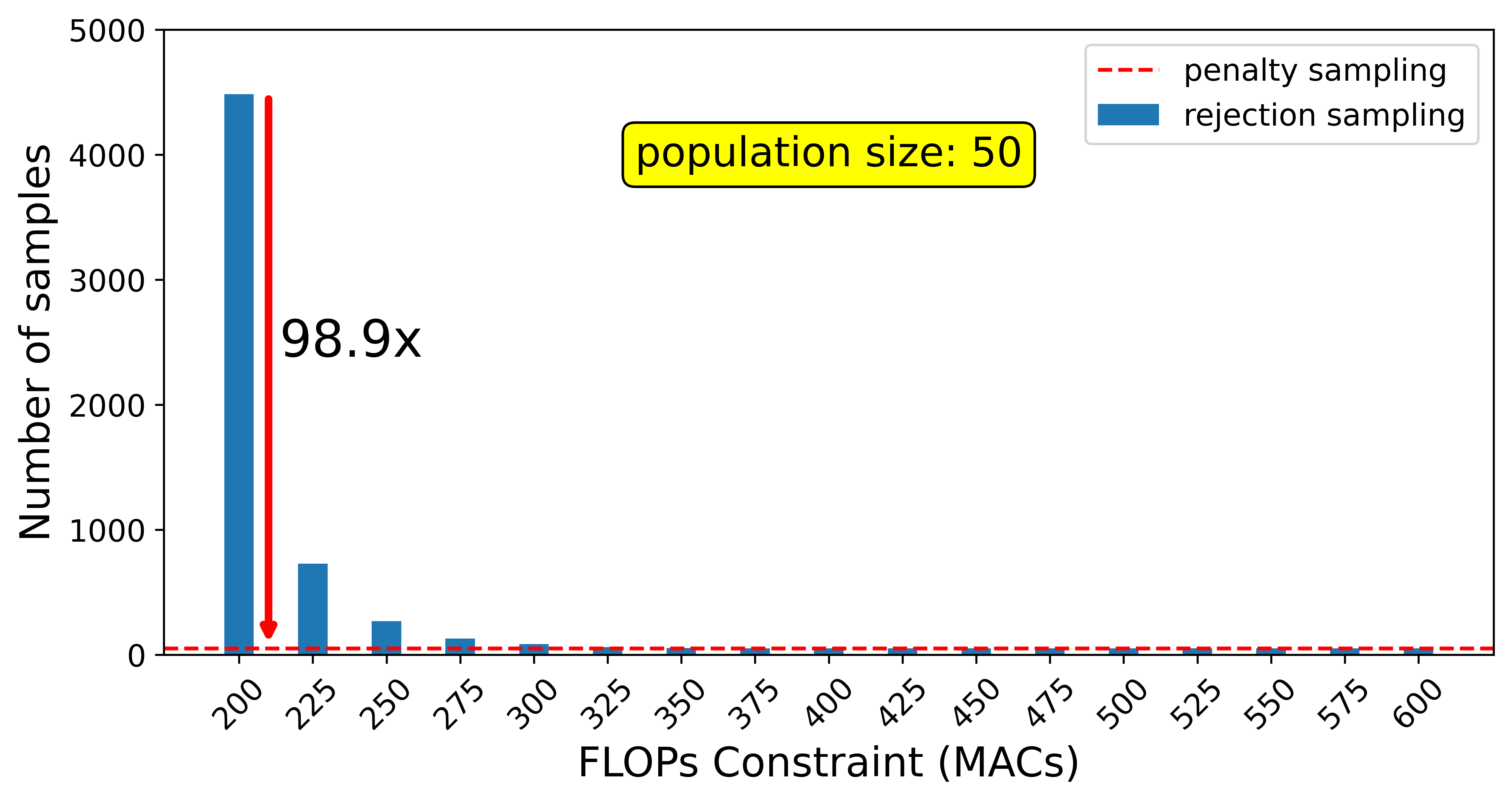}
    \end{center}
    \caption{Number of samples (x-axis) required to create a population of size 50 for various hardware constraints (y-axis). The reported number of samples required is the average of 10 runs.}
    \label{fig:rejection-sampling-vs-penalty}
\end{figure}

\comm{
\subsubsection{Without Penalty term}
    In Table~\ref{table:NAS201}, we present the results of \methodname$\ $in the NAS-Bench-201
    search space with no penalty term in Eq.(\ref{eq:fitness}). From the table, we can see that
    \methodname$\ $performs better than the previous evolution-based methods in terms of test and
    validation accuracy. This is mainly due to the effectiveness of RMI score as a performance
    estimator in Eq.(\ref{eq:fitness}).

\begin{table*}[t]
    \caption{Comparison of proposed method with NAS methods on NAS-Bench-201
    \cite{Dong2020NAS-Bench-201} with mean $\pm$ std. accuracies on CIFAR-10, CIFAR-100, ImageNet-16-120 from 10 runs each (higher is better). Search times are given for a CIFAR-10 search on a single GPU. Optimal refers to the best architecture accuracy for each dataset. The state-of-the-art results are adapted from \cite{zheng2022neural}.}
    \label{table:NAS201}
    \centering
    \scalebox{0.79}{
    \begin{tabular}{l|c|cc|cc|cc|c}
    \hline
     \bf{Method}& \bf{Search} & \multicolumn{2}{c|}{\bf{CIFAR-10}} & \multicolumn{2}{c|}{\bf{CIFAR-100}} & \multicolumn{2}{c|}{\bf{ImageNet-16-120}} & \bf{Search}\\
    &(seconds)& \it{validation} & \it{test} & \it{validation} & \it{test} & \it{validation} & \it{test}
    & \bf{Method}\\
    \hline
    
    RSPS\comm{$^\dagger$} \cite{li2019random} & $7587.12$ &$84.16\pm1.69$&$87.66\pm1.69$&$59.00\pm4.60$&$58.33\pm4.64$& $31.56\pm3.28$&$31.14\pm3.88$& random\\
    
    DARTS-V1 \cite{liu2018darts2} &$10889.87$ & $39.77\pm0.00$ & $54.30\pm0.00$ & $15.03\pm0.00$ &$15.61\pm0.00$ & $16.43\pm0.00$&$16.32\pm0.00$& gradient-based\\
    
    DARTS-V2 \cite{liu2018darts2} & $29901.67$ & $39.77\pm0.00$ & $54.30\pm0.00$ & $15.03\pm0.00$ &$15.61\pm0.00$& $16.43\pm0.00$&$16.32\pm0.00$& gradient-based\\
    
    GDAS\comm{$^\dagger$} \cite{dong2019searching} & $28925.91$ & $90.00\pm0.21$ & $93.51\pm0.13$ & $71.14\pm0.27$ &$70.61\pm0.26$& $41.70\pm1.26$&$41.84\pm0.90$& gradient-based\\
    
    SETN\comm{$^\dagger$}\cite{dong2019one} & $31009.81$ &$82.25\pm5.17$&$86.19\pm4.63$&$56.86\pm7.59$&$56.87\pm7.77$ &$32.54\pm3.63$&$31.90\pm4.07$& gradient-based\\
    
    ENAS\comm{$^\dagger$} \cite{pmlr-v80-pham18a} & $13314.51$ &$39.77\pm$0.00&$54.30\pm0.00$&$15.03\pm00$&$15.61\pm0.00$& $16.43\pm0.00$&$16.32\pm0.00$& RL\\
    
    RMI-NAS \cite{zheng2022neural} & \bf{1258.21} &\bf{91.44}$\pm$\bf{0.09} & \bf{94.28}$\pm$\bf{0.10} & \bf{73.38}$\pm$\bf{0.14} & \bf{73.36}$\pm$\bf{0.19} & \bf{46.37}$\pm$\bf{0.00} & 46.34$\pm$0.00 & random forrest\\

    \hline

    EvNAS \cite{sinha2021evolving} & 22444.78 &88.98$\pm$1.40 & 92.18$\pm$1.11 & 66.35$\pm$2.59 & 66.74$\pm$3.08 & 39.61$\pm$0.72 & 39.00$\pm$0.44 & evolution\\

    pEvoNAS& 4509 & 90.54$\pm$0.57 & 93.63$\pm$0.42 & 69.28$\pm$2.13 &69.05$\pm$1.99 & 40.00$\pm$3.22 & 39.98$\pm$3.76& evolution\\

    AmoebaNet\comm{REA} (\cite{real2019regularized}) & 12000 & $91.19\pm0.31$ & $93.92\pm0.30$ & $71.81\pm1.12$ &$71.84\pm0.99$ & $45.15\pm0.89$&$45.54\pm1.03$& evolution\\

    \bf{\methodname} (Our) & 6120 &\bf{91.44}$\pm$\bf{0.11} & 94.24$\pm$0.13 & 72.65$\pm$1.13 & 72.77$\pm$1.01 & 45.89$\pm$0.39 & \bf{46.40}$\pm$\bf{0.15} & evolution\\
    
    \hline
    Optimal & N/A &$91.61$&$94.37$&$73.49$&$73.51$&$46.77$&$47.31$& N/A\\
    
    \end{tabular}
    }
\end{table*}
}

\vspace{-0.17in}
\subsubsection{Penalty Term vs Rejection Sampling}
\label{subsubsection:penalty}
    In \textit{rejection sampling}~\cite{cai2019once, zheng2022neural}, the search method rejects the sampled architecture from the search space if it does not satisfy the hardware constraint.
    This leads to situations where the smaller the sub-space that satisfies that constraint, the more costly the sampling process becomes, see Figure~\ref{fig:rejection-sampling-vs-penalty}.
    This is due to the high probability of sampling architectures from a bigger sub-space that do not satisfy the hardware constraint.
    Figure~\ref{fig:rejection-sampling-vs-penalty} plots the average number of samples required to create a population of architectures with size 50 under different FLOPs constraints in 10 runs.
    For rejection sampling, Figure~\ref{fig:rejection-sampling-vs-penalty} shows that as we reduce the hardware cost constraint (in MACs), the number of samples increases exponentially reaching 4500 samples required for the 200 MACs constraint.
    In contrast, our approach requires a constant number of samples (i.e. 50) for all hardware constraints (a 98.9\% reduction in the number of samples for 200 MACs constraint). It does not reject any sampled architecture that does not satisfy the hardware constraint. However, it uses the penalty term, Eq.(\ref{eq:penalty}), to reduce the fitness, Eq.(\ref{eq:fitness}), of the sampled architecture that does not satisfy the hardware constraint.

\vspace{-0.1in}
\subsubsection{Is FLOPs Constraint a Good Indicator for Hardware Cost?}
\label{subsubsection:flopsgood}
    Figure~\ref{fig:exp_all_edge_devices} (h) shows architectures found using FLOPs (MACs) as a hardware cost proxy constraint against their respective hardware cost (i.e. latency) for different edge devices on CIFAR10.
    It is obvious that FLOPs constraint is not a good indicator of hardware cost as the same architecture will have different latencies in different edge devices.
    Furthermore, architectures with higher FLOPs values have lower latencies on some devices as compared to the latencies on other devices with lower FLOPs.
    For example, architectures with 200 MACs as FLOPs constraint have lower latency on pixel3 as compared to the latency on raspberry pi 4 (raspi4) of the discovered architectures with 80 MACs as the FLOPs constraint.
    Note that the reduction in the search cost of \methodname$\ $in Table~\ref{table:OFA} is mainly attributed to the reduction of the performance estimation cost.
    The search cost of all methods in Table~\ref{table:OFA} will increase if hardware latency is used as hardware cost.
    In our method, we primarily focus on the reduction of the performance estimation cost as it is the bottleneck part of any HW-NAS method.
\comm{
\section{Discussion}
\label{section:discussion}
    The use of RMI score for estimating the performance of an architecture makes \methodname$\ $
    dependent on the choice of the reference model used for calculating the RMI score
    (Section~\ref{subsubsection:RMIeffect}). An interesting future direction will be to perform
    a architecture search for GPU and then use \methodname$\ $to find the architecture for the edge
    devices with the discovered architecture for GPU as a reference model.
}

\vspace{-0.1in}
\section{Conclusion and Discussion}
\label{section:conclusion}
    
    In this work, we propose an efficient HW-NAS search algorithm that rephrases the NAS problem as a process of finding similar performing architecture with respect to a reference model for a target hardware.
    This is achieved by utilizing a DNN layer representation similarity metric, RMI score, as a proxy to evaluate architectures performance. A penalty term to penalize the search process is used. It controls the proportion of how far a hardware cost of an architecture is from the given hardware constraint on a target device.
    The proposed method is evaluated on two different search spaces. It showed a significantly lower search time that resulted into speedups of up to 8000$\times$. This resulted directly into lower usage of computational resources and lower $CO_2$ emissions consequently.
    Furthermore, the robustness of the proposed method is demonstrated on finding high performing architectures for the classification task on six different edge devices using two different types of hardware cost metrics (FLOPs and hardware latency).
    The use of RMI score for estimating the performance of an architecture makes \methodname$\ $
    dependent on the choice of the reference model used for calculating the RMI score
    (Section~\ref{subsubsection:RMIeffect}).
    An interesting future direction will be to investigate the effect of choosing different reference models.
    The effectiveness of \methodname$\ $for classification task encourages to extend current work to more downstream tasks such as object detection, image segmentation etc.

\vspace{-0.1in}
\section{Acknowledgement}
    This work is supported by the Luxembourg National Research Fund (FNR), under the project reference C21/IS/15965298/ELITE.


\medskip

{\small
\bibliographystyle{ieee_fullname}
\bibliography{egbib}
}

\pagebreak

\onecolumn
\begin{center}
	\textbf{\large Supplementary}
\end{center}

\setcounter{equation}{0}
\setcounter{figure}{0}
\setcounter{table}{0}
\setcounter{page}{1}
\setcounter{section}{0}
\renewcommand\thesection{S\arabic{section}}
\renewcommand{\theequation}{S\arabic{equation}}
\renewcommand{\thefigure}{S\arabic{figure}}

\begin{figure*}[h]
	\centering
	\begin{center}
		\includegraphics[width=\linewidth]{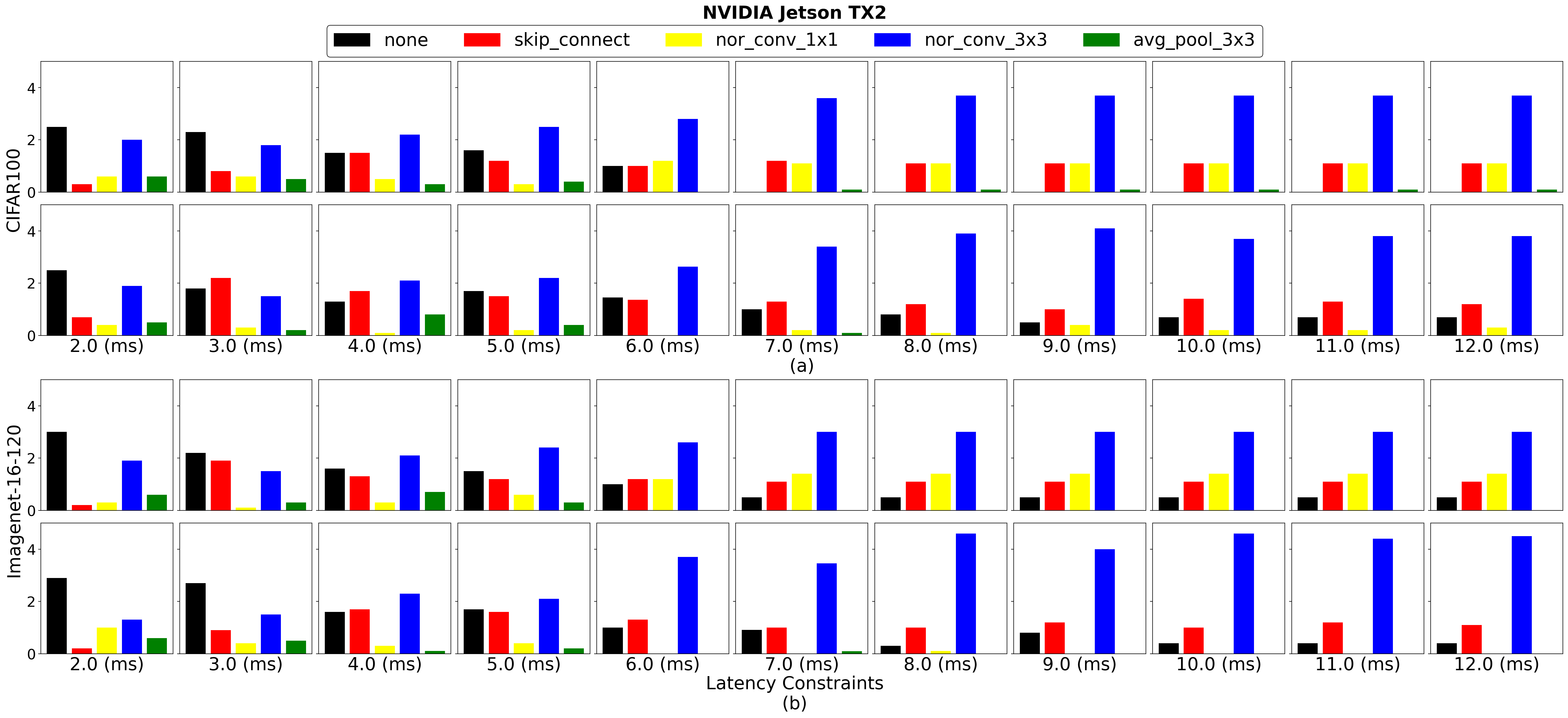}
	\end{center}
	\caption{
		Results of HW-NAS with different latency constraints for (a) CIFAR100 (b) ImageNet16-120 experiments.
		The top row of each experiment
		shows the average number of different types of operations (i.e. convolution 3x3, convolution 1x1, max pooling 3x3, skip connect and none) present in the top-10 architectures of NAS-Bench-201 in terms of test accuracy.
		The bottom row of each experiment shows the same quantities present in the architectures discovered from ten independent architecture search runs.
		Note that the architecture search is performed in the NAS-Bench-201 search space for NVIDIA Jetson TX2 as the target hardware in this case.
		The similarity of the distribution of operations at different latency constraints may suggests that the proposed method converges to the high performing architecture sub-space.}
	\label{fig:c10_ops_dist_edgegpu_latency}
\end{figure*}

\begin{figure*}[t]
	\centering
	\begin{center}
		\includegraphics[width=\linewidth]{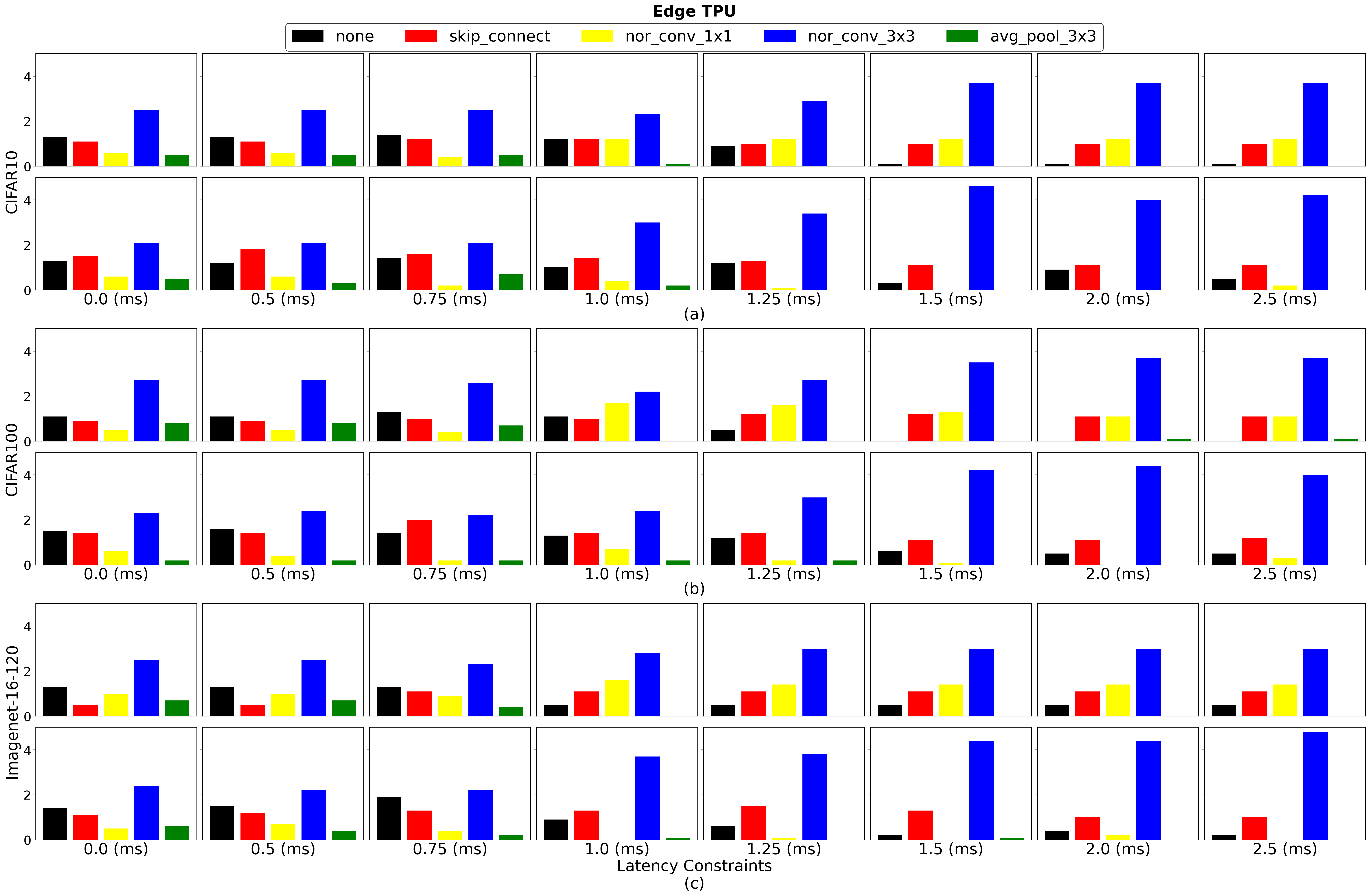}
	\end{center}
	\caption{
		Same experiments as in Figure~\ref{fig:c10_ops_dist_edgegpu_latency} with different latency constraints for Edge TPU on (a) CIFAR10 (b) CIFAR100 (c) ImageNet16-120.
	}
	\label{fig:ops_dist_edgetpu_latency}
\end{figure*}

\begin{figure*}[t]
	\centering
	\begin{center}
		\includegraphics[width=\linewidth]{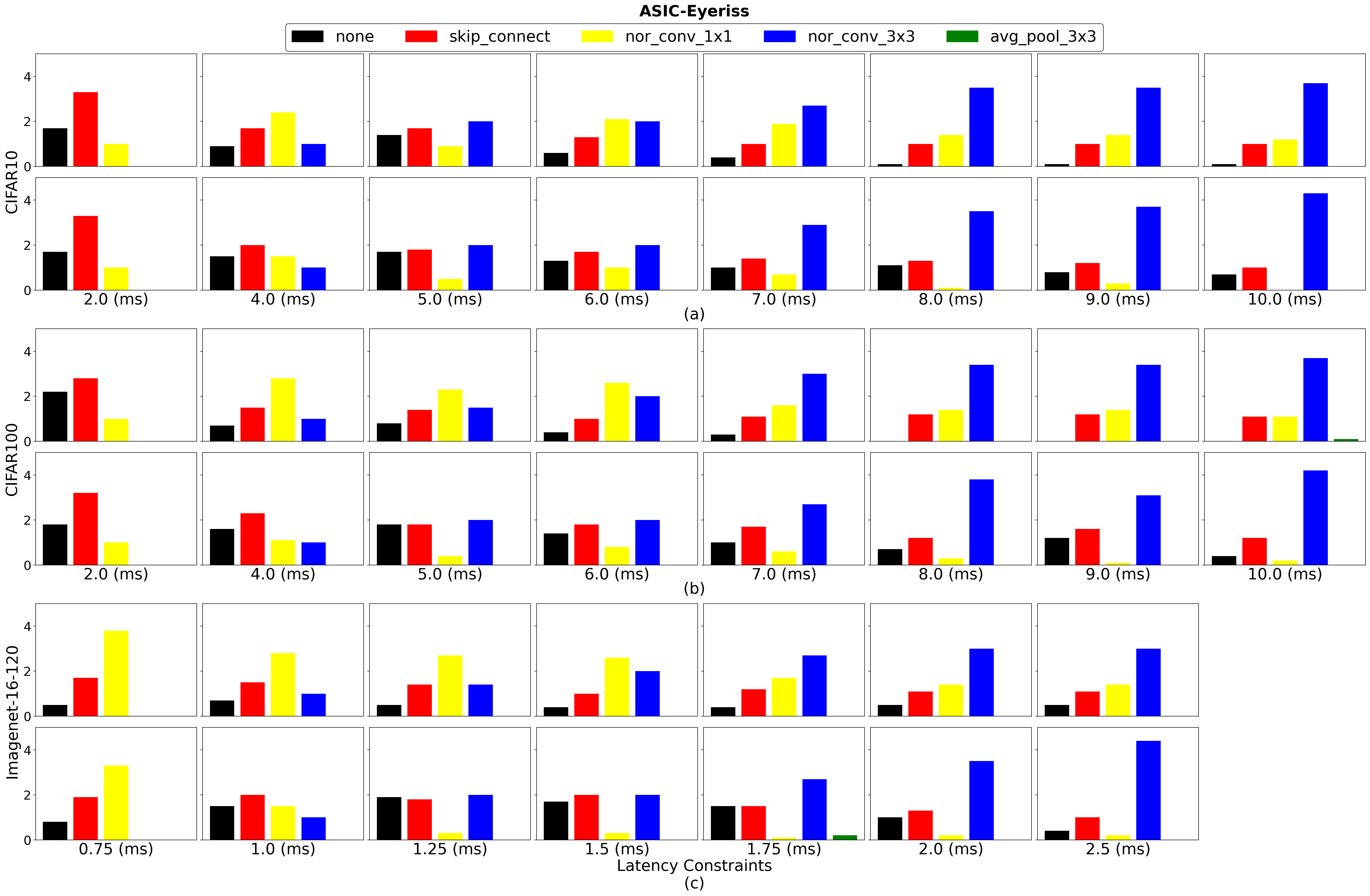}
	\end{center}
	\caption{
		Same experiments as in Figure~\ref{fig:c10_ops_dist_edgegpu_latency} with different latency constraints for ASIC-Eyeriss on (a) CIFAR10 (b) CIFAR100 (c) ImageNet16-120.
	}
	\label{fig:ops_dist_eyeriss_latency}
\end{figure*}

\begin{figure*}[t]
	\centering
	\begin{center}
		\includegraphics[width=\linewidth]{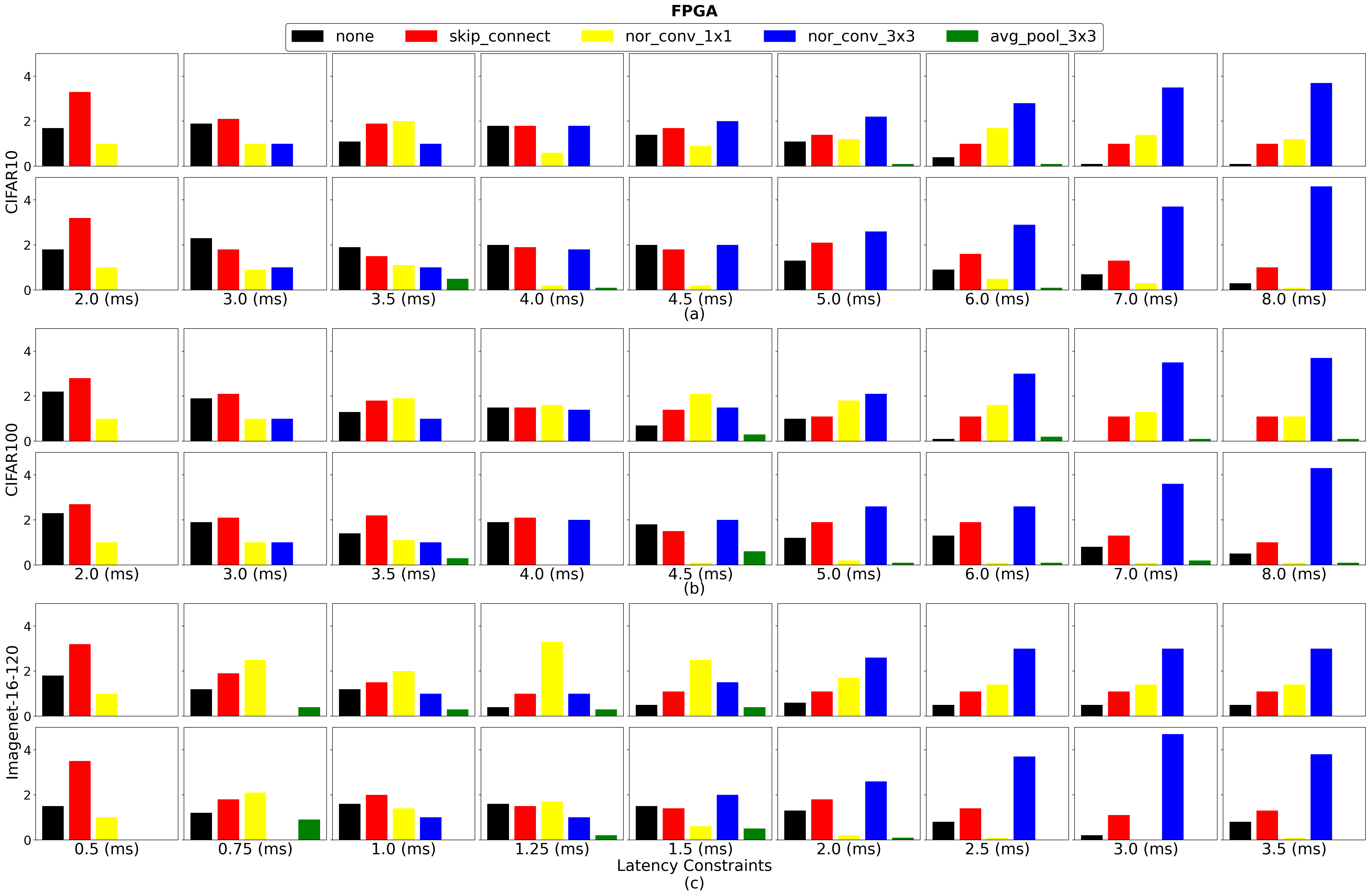}
	\end{center}
	\caption{
		Same experiments as in Figure~\ref{fig:c10_ops_dist_edgegpu_latency} with different latency constraints for FPGA on (a) CIFAR10 (b) CIFAR100 (c) ImageNet16-120.}
	\label{fig:ops_dist_fpga_latency}
\end{figure*}

\begin{figure*}[t]
	\centering
	\begin{center}
		\includegraphics[width=\linewidth]{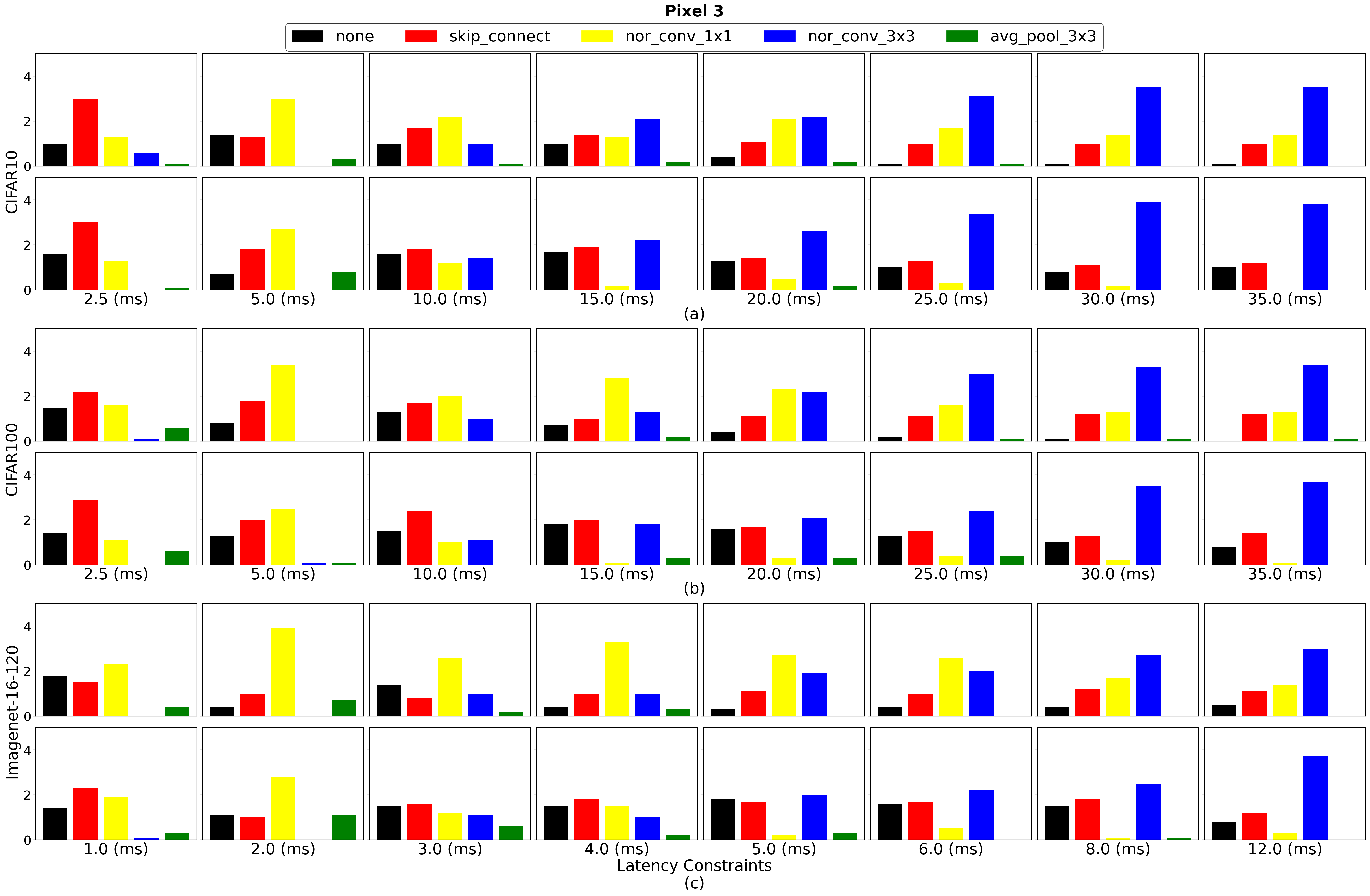}
	\end{center}
	\caption{
		Same experiments as in Figure~\ref{fig:c10_ops_dist_edgegpu_latency} with different latency constraints for Pixel 3 on (a) CIFAR10 (b) CIFAR100 (c) ImageNet16-120.
	}
	\label{fig:ops_dist_pixel3_latency}
\end{figure*}

\begin{figure*}[t]
	\centering
	\begin{center}
		\includegraphics[width=\linewidth]{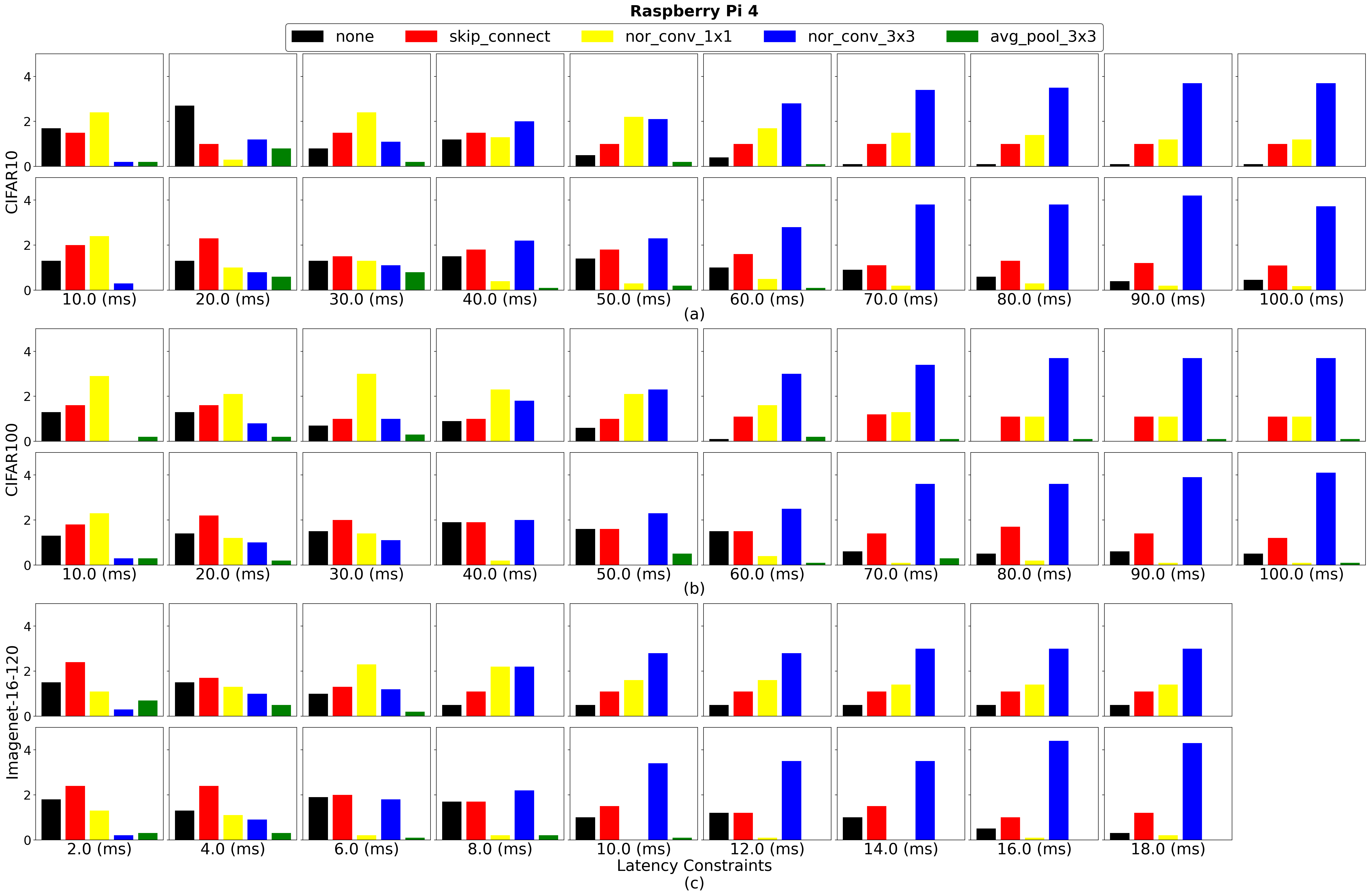}
	\end{center}
	\caption{
		Same experiments as in Figure~\ref{fig:c10_ops_dist_edgegpu_latency} with different latency constraints for Raspberry Pi 4 on (a) CIFAR10 (b) CIFAR100 (c) ImageNet16-120.
	}
	\label{fig:ops_dist_raspi4_latency}
\end{figure*}
\clearpage
\twocolumn

\section{Datasets}
\label{subsection:datasets}
\textbf{CIFAR-10} and \textbf{CIFAR-100} \cite{krizhevsky2009learning} have 50K training images and 10K testing images with images classified into 10 and 100 classes respectively.
\textbf{ImageNet}\cite{imagenet_cvpr09} is a well known benchmark for image classification containing 1K classes with 1.28 million training images and 50K test images.
\textbf{ImageNet-16-120} \cite{chrabaszcz2017downsampled} is a variant of ImageNet which is downsampled to 16x16 pixels with labels $\in\left[0,120\right]$ to construct ImageNet-16-120 dataset.

\end{document}